\def\year{2022}\relax
\documentclass[letterpaper]{article} 
\usepackage{aaai22}  
\usepackage{times}  
\usepackage{helvet}  
\usepackage{courier}  
\usepackage[hyphens]{url}  
\usepackage{graphicx} 
\usepackage{natbib}  
\usepackage{caption} 
\DeclareCaptionStyle{ruled}{labelfont=normalfont,labelsep=colon,strut=off} 
\usepackage{algorithm}
\usepackage{cite}
\usepackage{amsmath,amssymb,amsfonts}
\usepackage{textcomp}
\usepackage{xcolor}
\usepackage{booktabs}
\usepackage{leftidx}
\usepackage{subeqnarray}
\usepackage{cases}
\usepackage{textcomp}
\usepackage{epstopdf}
\usepackage{array} 
\usepackage{bm}
\usepackage{algpseudocode}
\usepackage{multirow}
\usepackage{newfloat}
\usepackage{listings}
\floatstyle{ruled}
\newfloat{listing}{tb}{lst}{}
\floatname{listing}{Listing}
\title{Social-DualCVAE: Multimodal Trajectory Forecasting Based on Social Interactions Pattern Aware and Dual Conditional Variational Auto-Encoder}

\author{
    Jiashi Gao,$^1$
    Xinming Shi,$^{1,2}$
    James J.Q. Yu$^1$
}
\affiliations{
    \textsuperscript{\rm 1}Department of Computer Science and Engineering, Southern University of Science and Technology, Shenzhen, China\\
    \textsuperscript{\rm 2}School of Computer Science, University of Birmingham, Birmingham, UK\\
    12131101@mail.sustech.edu.cn, xxs972@student.bham.ac.uk, yujq3@sustech.edu.cn

}

\usepackage{bibentry}

\begin{document}

\maketitle

\begin{abstract}
Pedestrians' trajectory forecasting is a fundamental task in multiple utility areas, such as self-driving, autonomous robots, and  surveillance  systems. The future trajectory forecasting is multi-modal, influenced by physical interaction with scene contexts and intricate social interactions among pedestrians. The mainly existing literature learns representations of social interactions by deep learning networks, while the explicit interaction patterns are not utilized. Different interaction patterns, such as following or collision avoiding, will generate different trends of next movement, thus, the model’s awareness of social interaction patterns is important for trajectory forecasting. Moreover, the social interaction patterns are privacy concerned or lack of labels. To jointly address the above issues, we present a social-dual conditional variational autoencoder (Social-DualCVAE) for multi-modal trajectory forecasting, which is based on a generative model conditioned not only on the past trajectories but also the unsupervised classification of interaction patterns. After generating the category distribution of the unlabeled social interaction patterns, DualCVAE, conditioned on the past trajectories and social interaction pattern, is proposed for multi-modal trajectory prediction by latent variables estimating. A variational bound is derived as the minimization objective during training. The proposed model is evaluated on widely used trajectory benchmarks and outperforms the prior state-of-the-art methods.
\end{abstract}
\section{Introduction}
\label{sec1}
Pedestrians' trajectory forecasting is a highly critical issue in intelligent systems, for its large-scale applicability from transportation systems to surveillance systems. Anticipating pedestrian movement is essential for collision avoidance and accident reduction in autonomous vehicles. In urban facilities, the surveillance systems detect and track any suspicious pedestrian's activities depending on the accurate trajectory prediction from the video. Pedestrian's behavior is influenced by two main factors, the physical scene around them and the social interactions with others. However, the performance of pedestrian trajectory prediction is hindered by several challenges:
\begin{itemize}
    \item Physical scene interaction: Physical scene interactions commonly exist in open scenes. The physical scenes contain multiple static facilities,  which may hinder and change the movement of pedestrians, such as buildings, impassable roads, and trees, etc.
    \item Social interaction: Compared with physical scene interaction, social interactions are more intricate for dynamic, diversity, and privacy. People socially interact with other pedestrians by multiple patterns such as follower-leader, collision avoidance, and group, etc.
    \item Multimodality: Different from the single-mode prediction problem, in pedestrian trajectory prediction, each agent has an amount of plausible trajectories to the same goal.
\end{itemize}

Deep learning \cite{wk:14,kt:12,bk:16}, with "context-aware", had been applied to learn the physical interactions. In the mainstream methods, the scene understanding tasks are regarded as generating captions for a scene image. Embedding layers, such as convolution networks, are unitized for scene feature extraction before trajectory forecasting.

\begin{figure}
    \centering
    \includegraphics[width=6cm]{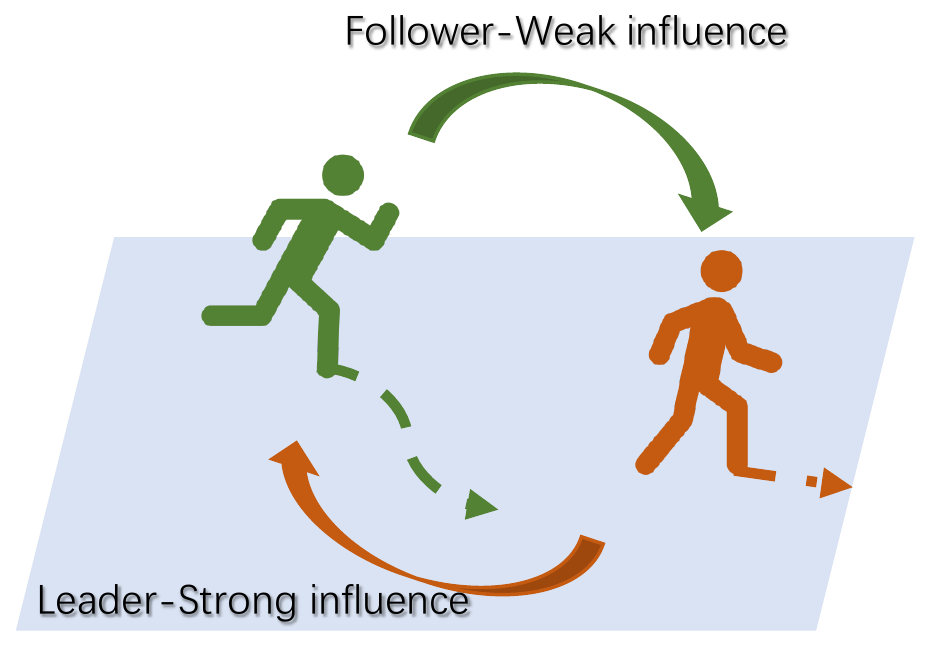}
    \caption{Different influence between two pedestrians, orange pedestrian's movement has more influence on green pedestrian's movement. }
    \label{fig1}
\end{figure}

In social interaction modeling, traditional approaches use hand-crafted functions such as social forces to represent the social interaction strength, but are limited to capture simple and neighboring interactions \cite{hb:95}. Recent state-of-the-art works \cite{ah:16,gp:18,l:17} learn to model social interactions in a data-driven way . For these methods, aggregation mechanisms are usually based on invariant symmetric functions, such as max-pooling or ordering functions sorting by euclidean distance, failing to accurately represent the interactions among pedestrians. Moreover, pedestrians in prior methods adjust their paths only by neighboring interactions without considering all people in the scene. To address the limitations of these works, Social-BiGAT  \cite{ks:19} captures social interactions by a graph representation. While the graph topology is intuitive to model social relations, Social-STGCNN \cite{mh:20} argued the social-BiGAT did not make the maximum utilization of graph.  Social-STGCNN designed a kernel function about the euclidean distance to distribute symmetric weights on edges, however, the influence weights need to be distinguished between two pedestrians in the real-world (Fig.~\ref{fig1}). Besides, the existed methods learn the social interactions implicitly, which embed the interactions in the feature vector of each agent. We argue that the awareness of social interaction pattern would improve the trajectory forecasting ability.

Drawing inspiration from recent developments in generative model, we propose Social-DualCVAE to learn multi-modal trajectory distributions under intricate social interactions, which is based on a conditional variational autoencoder (CVAE) \cite{sh:15}. Considering the issues of privacy and lacking of labels, an unsupervised classification is firstly utilized to explicit the social interaction patterns as the prior knowledge for the further prediction. First, the social interaction pattern of each pedestrian to another pedestrian is identified, which addresses the issue of undistinguished interactions between two pedestrians in prior work \cite{mh:20}. Next, we encourage multimodal trajectory distribution by a generative model, DualCVAE, constructing the mapping between the predicted trajectories and latent variables that represent the pedestrian movement intent in a scene. The proposed model allows us to generate trajectories that are socially plausible, while is also pattern aware by conditioning on the past trajectories and the categorical distribution generated by our proposed unsupervised classification.  Finally, we derived the empirical lower bound of the proposed DualCVAE as training loss.

\section{Related Work}
\label{sec2}
\subsubsection{Research On Social Interaction Model}
 Social force (SF) \cite{hb:95} is early used for human-human interactions by modeling interactions through forces (repulsion or attraction), which is extensively used in crowd analysis \cite{hb:05}, target tracking \cite{pg:10,kp:15} and robotics \cite{fr:13}. However, the SF models fail to generalize properly for hand-crafted features based. Recent approaches mainly model social interaction by deep neural networks, which capture human interaction features directly from data. Long short term memory networks (LSTM) \cite{hc:97}, a variant structure of recurrent neural networks (RNN), is widely used. Social LSTM \cite{ah:16} designs a social pooling layer, sharing hidden state information from LSTMs with the neighbouring pedestrians, to capture typical interactions even occurring in a distant future. Social LSTM ignores the important current intention of the neighbors. SR-LSTM \cite{z:19} and its extends reweighs the contribution of neighboring pedestrian to target via a social-aware information selection mechanism. Another typical reweighing mechanism is attention mechanism. Social-BiGAT \cite{ks:19} formulated the  pedestrian social interactions as a graph and distributed higher edge weights to more important interactions. Arguing that modeling the time and social dimensions separately could lead to sub-optimal solution, Agent-Former \cite{y} learns the representations of social interaction from both the time and social dimensions. The mentioned works pays more attention on feature extraction and representation learning, and fails to learn the interaction patterns explicitly. 
\subsubsection{Research On  Multimodal Trajectory Forecasting}  As the future trajectory of a pedestrian is uncertain and  multi-modal, several articles \cite{ah:16,mh:20}  modeled the pedestrian trajectories as a bi-variate Gaussian distribution. Recently, deep generative model, such as  generative adversarial networks (GANs) \cite{gp:18,lc:19,sd:19} and CVAE \cite{sh:15}, are used for multi-modal trajectory forecasting. Social-BiGAT \cite{ks:19}, following Bicycle-GAN, realized multi-modal prediction by constructing a bijection between the outputted trajectories and the latent noise vector, at the expense of a multi-step training process. Another direction is CVAE.  DROGON \cite{ch:19} utilized spatial-temporal graphs combined with CVAE, where the conditions  are represented by estimated intentions and history trajectory. Bitrap \cite{yao:21}, different from intentions estimating, estimates the goal of pedestrians and introduces a bi-directional decoder to improve the ability for longer-term  trajectory forecasting.

\section{Preliminaries}
\label{sec3}
We first clarify the related terms in pedestrian trajectory prediction problem and then introduce CVAE, a deep generative model that will be used throughout our work.
\subsection{Problem Definition}

Formally defined, the pedestrians trajectory forecasting is to generate the future movements of $T_{pred}$ steps, given the past observed movements of $T_{obs}$ steps. We denoted the prior observed trajectories of $N$ pedestrians as $\mathbf{X} =\left \{ x_{1},...,x_{N}  \right \} $. The trajectory of each pedestrian is a joint-state sequence of $pos_x$ and $pos_y$ coordinates in the 2D space, noted as $x_{i}= \left \{ (pos_{xi}^{t},pos_{yi}^{t}) \in \mathbb{R} ^{2} | t=1,...,T_{obs} \right \}$ for $\forall  i\in \left \{ 1,...,N \right \} $. The ground truth of the future trajectories is  $\mathbf{Y} =\left \{ y_{1},...,y_{N}  \right \} $ is over $T_{pred}$ timesteps and each pedestrian's future trajectory is noted as $y_{i}= \left \{ (pos_{xi}^{t},pos_{yi}^{t}) \in \mathbb{R} ^{2} | t=1,...,T_{pred} \right \}$ for $\forall  i\in \left \{ 1,...,N \right \} $. Different from former multi-modal prediction works, which set the goal to learn the feasible future trajectories $\hat{Y}$ from a conditional distribution $p(\mathbf{\hat{Y}} |\mathbf{X})$, we aim to generate trajecory samples from $p(\mathbf{\hat{Y}} |\mathbf{X},\mathbf{C})$  conditioned on  the past movements $\mathbf{X}$ and the social interaction category $\mathbf{C}$. 
\begin{figure*}[h]
    \centering
    \includegraphics[scale=0.25]{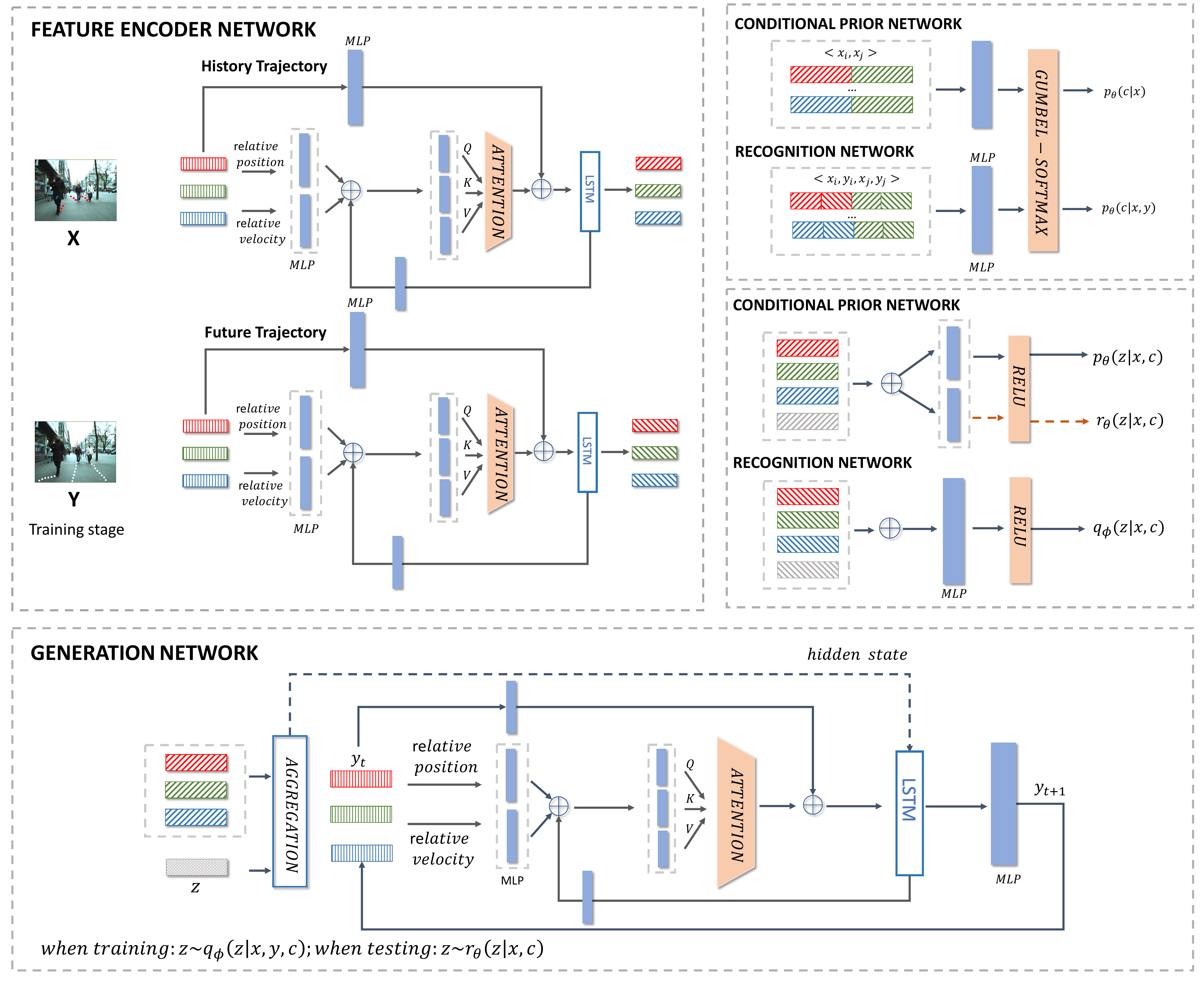}
    \caption{Architecture for the proposed Social-DualCVAE model, consisting of two feature encoder networks, two conditional prior networks, two recognition networks and a generation network.}
    \label{fig2}
\end{figure*}
There is no clear idea  the exact number of all possible social interaction types. Though the existed research \cite{kt:21} categorized social interactions to four main types: static, linear, interacting and non-interacting, and the interacting type can be further divided to four sub-categories, leader-follower, collision avoidance, group and other interactions. It is unknown whether these categories  cover all existing social interactions, and whether such classification is suitable for prediction. In our work, we set the number of possible social interaction types to be a hyperparameter $\mathcal{H}$. The social interaction category  $\mathbf{c}$ is sampled following $
p(\mathbf{C} |\mathbf{X})$, which is another training goal in our work. Note that the physical scene are not considered here, focusing on the effect of social interaction modeling,.

\subsection{Deep Generative Models}
Let $\mathbf{D} =\left \{\mathbf{X},\mathbf{Y} \right \} $ be a training dataset of history trajectories $\mathbf{X}$ and future trajectories $\mathbf{Y}$, from a statistical perspective, the goal of multi-modal human trajectory prediction is to draw a data distribution $p(y|x)$ about target data $y\in \mathbf{Y} $, where $x\in \mathbf{X} $ is known conditions. Introduced a latent variables $z $, the predictive distribution $p(y|x)$ in relation with $z$ can be reparameterized as:
\begin{equation}
    p(y|x)=\int p(y|x,z)p(z)dz.
\end{equation}
The generator in generative model is a deterministic function aiming to map the samples of $z\sim p(z)$ to the samples of $y\sim p(y|x,z)$. This function could be implemented by a deep neural network $G_{\theta}(x,z)$ with parameter $\theta$. Thus, the generate process of $y$ can be described as:
\begin{equation}
    y=G_{\theta } (x,z).
\end{equation}

The $G_{\theta}(x,z)$ could be regarded as a sampling generator of distribution $p_{\theta}(y|x)$, thus the goal of generative models is to maximize  the  conditional log-likelihood $\mathrm{log}\, \,p_{\theta } (y|x)$. The issue is how to get unknown $p(z)$. In CVAE, one of various generative models,  $p(z)$ is inferred by $p_{\theta }(z|x)$  making the latent variables statistically independent of input variables. However, the integration with respect to $z$ is typically intractable as the real parameters of distribution $p_{\theta }(z|x)$ is not known yet. Here, we infer $p_{\theta }(z|x)$ through variational inference (VI), which minimize the Kullback-Leibler divergence (KLD)  between the prior distribution $p_{\theta }(z|x)$ with a variational distribution $q_{\phi }(z|x,y)$. By maximizing the conditional log-likelihood $\mathrm{log}\, \,p_{\theta } (y |x )$, the variational lower bound \cite{sh:15} of the CVAE model is derivated as follows:
\begin{equation}
\begin{aligned}
        &\mathrm{log}\, \,p_{\theta } (y |x )\ge -\mathcal{D}[q_{\phi}(z|x,y)||p_{\theta }(z|x)] \\&+\mathbb{E}_{q_{\phi}(z|x,y)}\left [  \mathrm{log}\, \,p_{\theta } (y |x,z )\right ],
\end{aligned}
\end{equation}
and the empirical lower bound (ELBO) is writtend as :
\begin{equation}
\begin{aligned}
        &\mathcal{L}_{elbo}(x ,y ;\theta ,\phi ) = -\mathcal{D}[q_{\phi } (z|x ,y )||p_{\theta } (z|x )] \\&+\mathbb{E}_{q_{\phi}(z|x,y)}\left [  \mathrm{log}\, \,p_{\theta } (y |x,z )\right ].
\end{aligned}
\end{equation}

\section{Proposed Framework}

 A high-level architectural view of the proposed Social-DualCVAE model is given in Fig.~\ref{fig2}. This model involves four network types: feature encoder networks to extract feature from raw trajectories, conditional prior networks $p_{\theta}(c|x)$, $p_{\theta}(z|x,c)$, and recognition networks $q_{\phi}(c|x,y)$, $q_{\phi}(c|x,y,c)$, to infer interaction pattern distribution and latent variable distribution, as well as a generation network $p_{\theta}(y|x,z)$ for trajectory forecasting. More details is introduced in the following. 
\subsection{Feature Encoder}
The feature encoders include two main modules, self-encoder and social-encoder. For self-encoder, we set the velocity (relative displacements) for each pedestrian calculated from raw coordination data as input. By a multi-layer perceptron (MLP), each velocity input $v$ is embedded to a higher dimension denoted as $h_{self}$. For the social-encoder, we first embed the relative distance $dist_{ij}$ and relative direction $dir_{ij}$ between $i$-th pedestrian and $j$-th pedestrian to $h^{dist}$ and $h^{dir}$ by MLP, respectively. The $h^{dist}$, $h^{dir}$ and the hidden state from LSTM $h^{lstm}$ are concatenated as  input vector of the multi-head attention network \cite{vs:17}, enhancing the ability of jointly attending to information from different representation sub-spaces of the input. The self-encoder outputs and social-encoder outputs are concatenated into LSTM, modeling features over time. The calculations of above are given by
\begin{equation}
\begin{aligned}
    \mathrm{self- encoder}:v_{i} =  \Delta x_{i},h^{self}=f_{self-encoder}(v). 
\end{aligned}
\end{equation}
\begin{equation}
\begin{aligned}
    &\mathrm{social-encoder}:dist_{ij} =  x_{j}-x_{i},dir_{ij} = v_{j}-v_{i}, \\& h^{dist} = f_{dist-encoder}(dist),h^{dir} = f_{dir-encoder}(dir),
    \\& Q=f_{query}(Concat(h^{dist},h^{dir},f_{lstm}(h^{lstm}))),
    \\& K=f_{key}(Concat(h^{dist},h^{dir},f_{lstm}(h^{lstm}))),
    \\&V_l=f_{value}(Concat(h^{dist},h^{dir},f_{lstm}(h^{lstm}))),
    \\&h^{social}= MultiHead(Q,K,V),
    \\& MultiHead(Q,K,V)=Concat(head_1,...,head_H)W^O,
\end{aligned}
\end{equation}
where
\begin{equation}
\begin{aligned}
head_l=\mathrm{softmax}   \left (  \frac{QW_{l}^{Q}(KW_{l}^{K})^{T}}{\sqrt{d_{k} } } \right ) VW_{l}^{V},
\end{aligned}
\end{equation}
\begin{equation}
\begin{aligned}
    h^{encoder}=LSTM(Concat( h^{self}, h^{social})).
\end{aligned}
\end{equation}

\subsection{Unsupervised Social Interaction Classification}
We model the distribution $p(c|x)$ of social interaction pattern category, an unsupervised multi-classification problem, by a MLP network $p_{\theta}(c|x)$ with parameter $\theta$. Due to the social interaction patterns are asymmetry, the feature vectors $h_i^{encoder}$ and $h_j^{encoder}$ of $i$-th pedestrian and $j$-th pedestrian, are concatenated orderly as input to classification networks. The dimension of output layer, equaling to the number of possible interaction types, is set to a hyperparameter $\mathcal{H}$. The classification network, also called conditional prior network, is expected to output the samples following conditional prior distribution $p_{\theta}(c|x)$. Stochastic neural networks with discrete random variables had been widely used to learn distributions encountered in unsupervised learning, however, stochastic networks with backpropagation algorithm, could not apply in non-differentiable layers directly. Gumbel-softmax \cite{j:16}, a stochastic gradient estimation whose parameter gradients can be easily computed via the reparameterization trick, is utilized to approximate categorical samples from a pattern distribution. This process could be written as
\begin{equation}
\begin{aligned}
     &h_{ij}^{pattern}=f_{pattern}(Concat(h_i^{encoder},h_j^{encoder})),\\& \forall i,j\in \left \{ 1,...,N \right \}\,\,\,\,and \,\,\,\,i\ne j
\end{aligned}
\end{equation}
\begin{equation}
p_{\theta }(c|x)=softmax(\frac{h^{pattern}+g}{\tau } ),
\end{equation}
where $g$ is sampled from $Gumbel(0,1)$ and $\tau$ is the temperature hyperparameter. Jang \textit{et al.} \cite{j:16} verified that the small $\tau$  would generate one-hot samples with  the larger variance of the gradients. On the contrary, the samples of the large $\tau$ are smooth but the variance of the gradient is small. 

Interaction patterns distribution is not only determined by the history trajectory, but also by the future motions. Thus, besides the conditional prior network $p_{\theta}(c|x)$, we built a recognition network $q_{\phi}(c'|x,y)$ conditioning on both history and future trajectory. The true conditional prior distribution $p_{\theta}(c|x)$ could be learned by minimizing the KLD  between $p_{\theta}(c|x)$ and $q_{\phi}(c'|x,y)$.

\subsection{Latent Variables Estimation}
In order to realize multi-modal trajectory prediction, we introduced a latent variable $z$ as the samples from latent intent distribution, which could also be regarded as the noise on an certain predicted trajectory. Assuming $z$ following an isotropic Gaussian distribution here, both  the conditional  prior network $p_{\theta}(z|x,c)\sim N(\mu ,\sigma ^{2} I)$ and recognition network  $q_{\phi}(z|x,y,c')\sim N(\mu' ,\sigma^{\prime 2} I)$ is constructed by MLP, outputing  mean $\mu$ and log variance $\sigma^2$ for each pedestrian.
\begin{equation}
    \mu=f_{\mu}(h^{encoder})
\end{equation}
\begin{equation}
    \mathrm{ log}\sigma ^{2} =f_{\sigma}(h^{encoder})
\end{equation}
Then  the reparametrization trick is used to obtain samples of $z$ from $q_{\phi}(z|x,y,c')$ during training or $p_{\theta}(z|x,c)$  during testing. This process could be written as
\begin{equation}
\begin{aligned}
    \epsilon \sim  \mathcal{N} (0,I), z=\mu +\epsilon \odot \sigma 
\end{aligned}
\end{equation}
\subsection{Multi-modal Prediction}
The generation network is designed to output displacement of each pedestrian one step at a time, the currently generated output is then feed-backed as input to produce the displacement of next time step. This autoregressive design mitigates compounding errors during testing time but with a higher expense of training speed \cite{y:21}. The architecture of generation network is similar to feature encoder network. 
\subsection{Losses}
When social interaction pattern $c'$ had sampled from the posterior distribution $q_{\phi}(c'|x,y)$ during training, the goal of variational inference was to minimize the KLD between the estimated posterior distribution $q(z|x,y,c')$ and conditional prior distribution $p(z|x,c') $, written as
\begin{equation}
\begin{aligned}
     &\mathcal{D}  [ q(z|x,y,c')||p(z|x,c')  ] =\\& E_{z\sim q(z|x,y,c')} [log\,\,q(z|x,y,c')  - log\,\,p(z|x,c') ],
\end{aligned}
\label{eq:14}
\end{equation}
Introduced an intermediate variable $log\,\,p(z|x,c)$, Eq.~\ref{eq:14} could be expanded by Bayes' rule, 


\begin{equation}
\begin{aligned}
    &\mathcal{D}  [ q(z|x,y,c')||p(z|x,c')  ] =E_{z\sim q(|x,y,c')}  [ log\,\,q(z|x,y,c')-\\&log\,\,p(z|x,c) +log\,\,p(z|x,c) - \\&  log\,\,p(y|z,x,c') -log\,\,p(z|x,c')  ] + log\,\,p(y|x,c').
\end{aligned}    
\end{equation}

\begin{figure}[h]
    \centering
    \includegraphics[width=8cm]{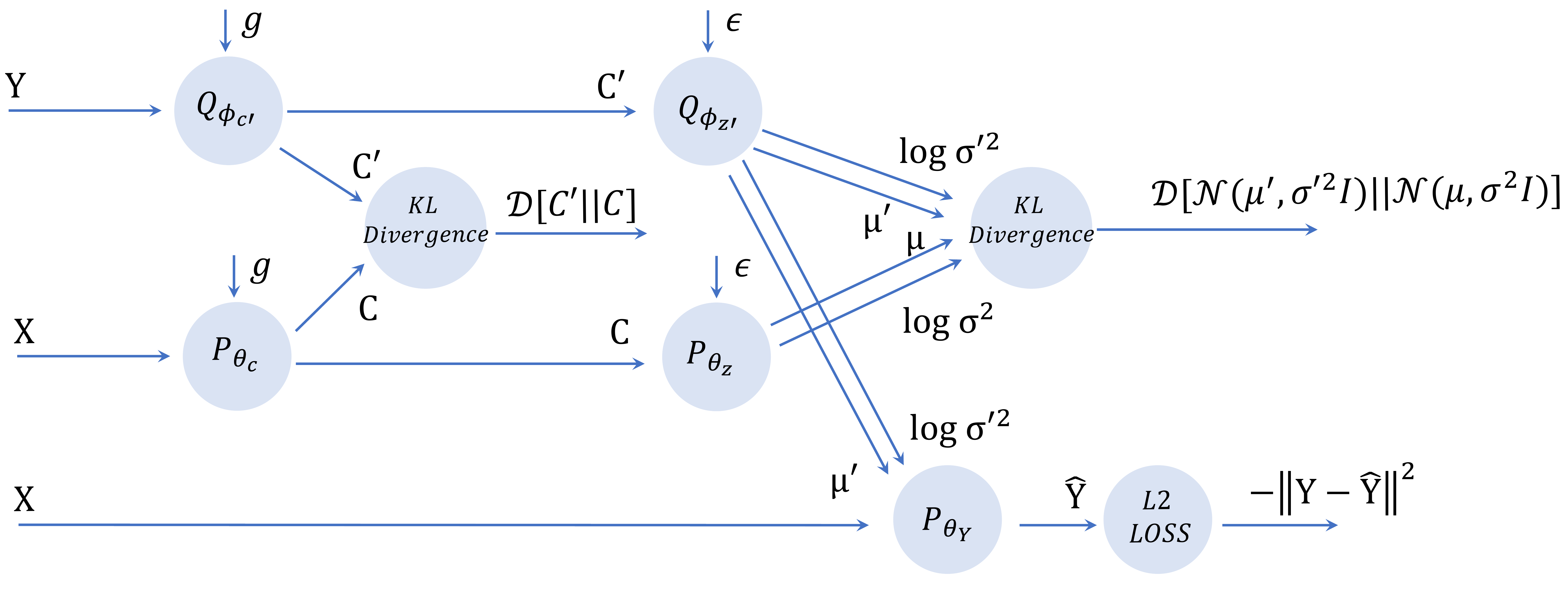}
    \caption{Data flow graph of Social-DualCVAE, where each node in the graph represents the instance of a mathematical operation or network module, and each edge is a multi-dimensional data (tensors) on which the operations are performed. }
    \label{fig3}
\end{figure}

Then, the variational lower bound  of the proposed DualCVAE model is derivated as 
\begin{equation}
\begin{aligned}
   & log\,\,p(y|x,c') \ge  E_{z\sim q(z|x,y,c')} \left [ log\,\,p(y|x,z,c')\right ] \\ &-  \mathcal{D}\left [ q(z|x,y,c')||p(z|x,c)\right ] \\ & -\mathcal{D}\left [ p(z|x,c)||p(z|x,c')\right ].
\end{aligned}    
\end{equation}
When $c'\sim q(c'|x,y)$ and $c\sim p(c|x)$, a special solution to minimize the KLD between $p(z|x,c')$ and $p(z|x,c)$ is to minimizing the KLD between $p(c|x)$ and $q(c'|x,y)$. The loss function could be derived as 
\begin{equation}
\begin{aligned}
 &\mathit{Loss} =-\left \| Y-\hat{Y}    \right \|_{2} +\lambda _{z} \mathcal{D}\left [ q(z|x,y,c')||p(z|x,c)\right ]\\& +\lambda _{c}\mathcal{D}\left [ q(c'|x,y)||p(c|x)\right ],
\end{aligned}    
\end{equation}
where, $\lambda_z$ and $\lambda_c$  are hyparameters to weigh the loss terms. When defining the latent distribution as Gaussian distribution $p(z|x,c)\sim N(\mu ,\sigma ^{2} I)$ and $q(z|x,y,c')\sim N(\mu' ,\sigma^{\prime 2} I)$, $\mathcal{D}\left [q(z|x,y,c')||p(z|x,c)\right ]$ could be rewritten as:
\begin{equation}
\begin{aligned}
    &\mathcal{D}[q(z|x,y,c')||p(z|x,c)]=\\& -\frac{1}{2} \sum \left [ log\frac{\sigma'^{2}  }{\sigma^{2}}  -\frac{\sigma'^{2}  }{\sigma^{2}}-\frac{(\mu'-\mu)^{2} }{\sigma^{2}} +1\right ].
\end{aligned}
\end{equation}
The  pattern distribution of social interactions $q(c'|x,y)$ and $p(c|x)$ is $softmax$ distribution, then
\begin{equation}
\begin{aligned}
   &\mathcal{D}[q(c'|x,y)||q(c|x)] =\\& H\left ( q(c'|x,y),p(c|x) \right ) -H\left ( q(c'|x,y),q(c'|x,y) \right ),
\end{aligned}
\end{equation}
where $H(q,p)$ is cross-entropy, denoted as 
\begin{equation}
\begin{aligned}
     H\left ( q, p \right ) =-\sum q\mathrm{log} (p).
\end{aligned}
\end{equation}
The data flow of Social-DualCVAE is shown in Fig.~\ref{fig3}.
\begin{figure*}[h]
    \centering
    \includegraphics[scale=0.55]{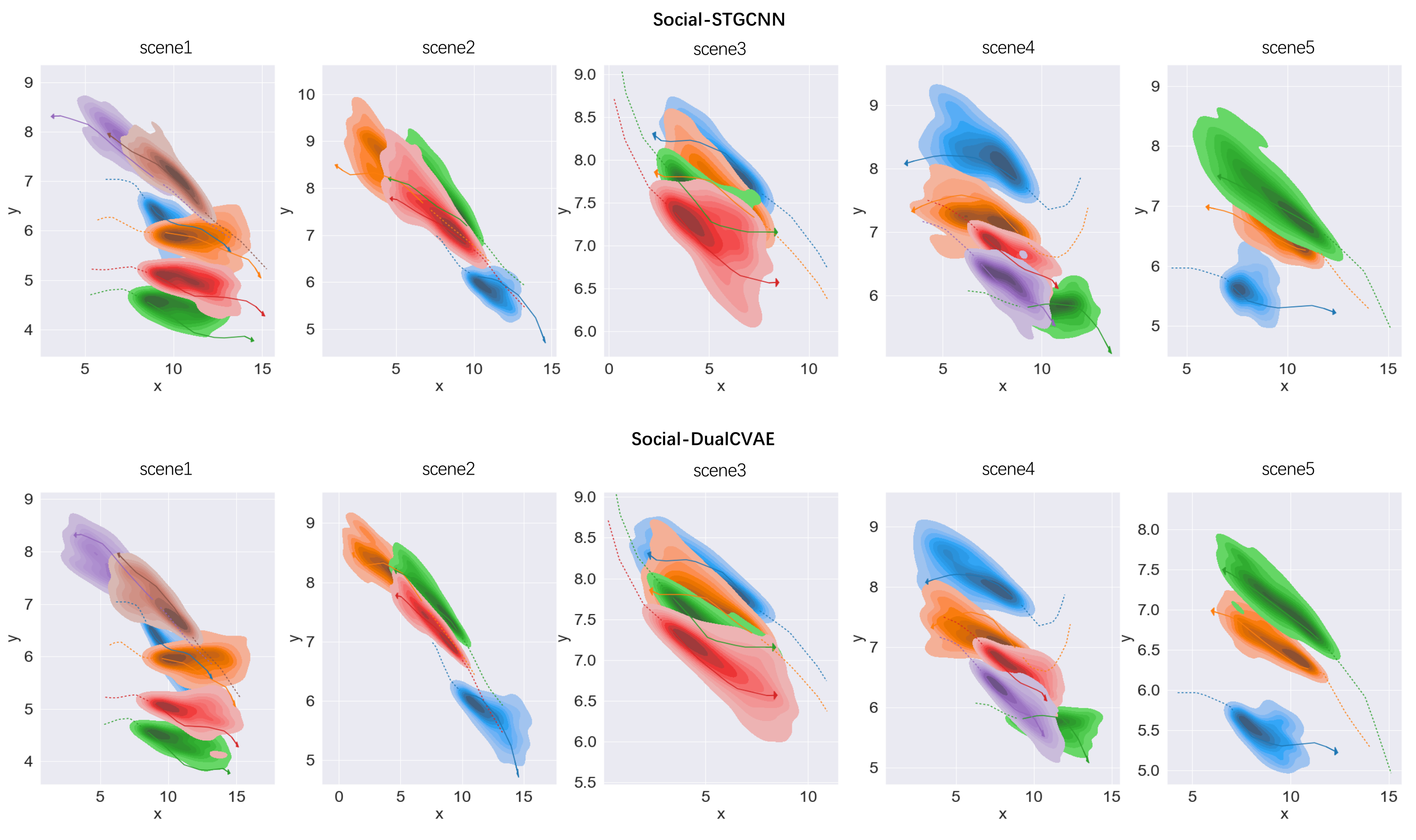}
    \caption{Qualitative analysis of Social-DualCVAE. Observed trajectories are denoted as solid lines, ground truth future movements are denoted as dashed lines, and generated multi-modal samples are demoted as contour maps. We compare models versus previous models, Social-STGCNN, in five scene frames from the ETH and UCY datasets. The scenes contains various of social interactions: walking in parallel or grouping (e.g. orange, red and green pedestrians in scene1), following or leading (e.g. orange and green pedestrians in scene2), meeting or avoiding collision (e.g. orange and green pedestrians in scene3) and no-interaction (e.g. blue and green pedestrians in scene5). }
    \label{fig4}
\end{figure*}


\begin{table*}[t]
\caption{min$ADE_{20}$ / min$FDE_{20}$ (Meters) of Trajectory Prediction}
\centering
\label{tab1}
\begin{tabular}{@{}ccccccc@{}}
\toprule
Method  & ETH         & HOTEL     & UNIV      & ZARA1     & ZARA2     & AVG       \\\cmidrule(l){1-1}\cmidrule(l){2-6}\cmidrule(l){7-7}
Linear             & 1.33 / 2.94 & 0.39/0.72 & 0.82/1.59 & 0.62/1.21 & 0.77/1.48 & 0.79/1.59 \\ 
Social-LSTM        & 1.09/2.35   &0.79/1.76           & 0.67/1.40         & 0.47/1.00          & 0.56/1.17          & 0.72/1.54 \\ 
Social-GAN         & 0.87/1.62   &0.67/1.37           &0.76/1.52           &0.35/0.68           & 0.42/0.84          & 0.61/1.21          \\ 
SoPhie             & 0.70/1.43   & 0.76/1.67          & 0.54/1.24          &0.30/0.63           &0.38/0.78           &0.54/1.15           \\ 
CGNS               & 0.62/1.40   &0.70/1.93           &0.48/1.22           &0.32/0.59           &0.36/0.71           &0.49/0.97           \\ 
PIF                & 0.73/1.65   & \textbf{0.30}/\textbf{0.59}          & 0.60/1.27          & 0.38/0.81          &0.31/0.68           & 0.46/1.00          \\ 
GAT                & 0.68/1.29   & 0.68/ 1.40          & 0.57/1.29          & 0.29/0.60          & 0.37/0.75          &0.52/1.07           \\ 
Social-BiGAT       & 0.69/1.29   & 0.49/1.01          & 0.55/1.32          &0.30/0.62           &0.36/0.75           &0.48/1.00           \\ 
Social-STGCNN      & \textbf{0.64}/\textbf{1.11}   &0.49/0.85           & 0.44/0.79          &0.34/0.53           &0.30/0.48           &0.44/0.75           \\ 
CVAE & 0.74/1.40   & 0.34/0.62 &0.40/0.80    &0.30/0.53           & 0.26/0.44          & 0.41/0.76           \\ 
Social-DualCVAE (2) & 0.68/1.24   & 0.35/0.65 &\textbf{0.39}/\textbf{0.72}    &0.31/0.54           & 0.29/0.48          & 0.40/0.73     \\ 
Social-DualCVAE (4) & 0.66/1.18   & 0.34/0.61 &\textbf{0.39}/0.74    &\textbf{0.27}/\textbf{0.48}           & \textbf{0.24}/\textbf{0.42} &\textbf{0.38}/\textbf{0.69}    \\ 
Social-DualCVAE (6) & 0.66/1.26   & 0.42/0.78 &0.40/0.77    &0.31/0.56           & 0.27/0.45          & 0.41/0.76     \\ \bottomrule
\end{tabular}
\end{table*}

\section{Experiments}
\subsubsection{Datasets}
We conduct expirements on two well-established datasets: ETH \cite{pl:09} and UCY \cite{lr:07}. Both contain annotated trajectories of multi-pedestrians in real world scenes, including rich social interactions. ETH dataset contains two scenes: ETH and HOTEL. UCY dataset contains three scenes: ZARA1, ZARA2 and UNIV. These five datasets are the major benchmarks for pedestrian trajectory prediction. Following the previous dataset segmentation\cite{ah:16}, the model is trained on a portion of a specific dataset, and tested on the rest. The other four datasets was used as validation set. All the trajectories in the datasets are sampled every 0.4 seconds. Evaluations occur over 8 seconds (20 timesteps), where the first 3.2 seconds (8 timesteps) is set as observed trajectory, and the model is trained to forecast the last 4.2 seconds (12 timesteps) correspond to predicted future trajectory.
\subsubsection{Baselines}
We compare our proposed Social-DualCVAE to several deterministic baselines, including a non-probabilistic method, Linear, minimizing least square error by a linear regressor, and three predictive models, Social-LSTM \cite{ah:16} with social pooling and LSTMs, Social-STGCNN \cite{mh:20} modeling the interactions as a graph and predicting sequence by TCNs, and PIF \cite{li:19} utilizing visual information from the behavioral sequence of the person, as well as five generative models: Social-GAN \cite{gp:18} which applies GANs to Social-LSTM, SoPhie \cite{sd:19}, which introduces  attention network to Social-GAN, CGNS \cite{lc:19}, which applies  variational divergence minimization for latent space learning, GAT, which applies graph attention network combined with GAN, and Social-BiGAT \cite{ks:19}, appending latent scene encoder on GAT. Moreover, we present evaluation results of four versions of our model, CVAE model with the similar architecture to Social-DualCVAE except social interaction pattern aware, and three Social-DualCVAEs with different hyparameter of social interaction categories.  
\subsubsection{Quantitative Metrics}
We utilize the average displacement error ($ADE_K$)  and final displacement error ($FDE_K$) to evaluate the performance of models. $ADE_K$ is defined as the  the minimum average displacement error along the predicted future trajectory over $K$ samples, while the $FDE_K$  is defined as the final displacement error of the end point over $K$ samples. A hold-one-out cross evaluation strategy \cite{gp:18,ks:19} is used to produce diverse samples  which generate $K$ possible outputs by randomly sampling $\epsilon$ from $\mathcal{N} (0,I)$ and choosing the “best” prediction as prediction result.
\begin{equation}
    ADE_{K}=\underset{k\in K}{min}\frac{\sum_{n\in N}^{}\sum_{t\in T_{pred}}^{}\left \| y_{n}^{t} -\hat{y}_{n}^{t}  \right \|_2  }{N\times T_{pred} } 
\end{equation}
\begin{equation}
    FDE_{K}=\underset{k\in K}{min}\frac{\sum_{n\in N}^{}\left \| y_{n}^{T_{pred}} -\hat{y}_{n}^{T_{pred}}  \right \|_2  }{N} 
\end{equation}
\subsubsection{Training Setup}
We set the training batch size to 128. The model was trained for 60 epochs using adam optimizer. The initial learning rate is 0.001, and changed to 0.0001 after 30 epochs. To avoid over-fitting, the L2 regularization is activate and the regularization parameter is set to 0.1. Meanwhile, the dropout probability $p_{drop}$ is set to 0.2 for all dropout layers. The hyperparameter $\mathcal{H}$ of interaction catergory num is set to 2,4 and 6, respectively. The temperature hyperparameter $\tau$ is set to be 0.1. The weights hyperparameter $\lambda_z$ and  $\lambda_c$ in loss functions are set to be 0.005.
\subsection{Quantitative Results}
We compare our model to the baselines in Table 1, showing the $ADE_K$ and $FDE_K$ under different scenes. From overall, we see that the proposed Social-DualCVAE, setting total pattern categories to 4, shows the best performance. The $ADE_K$ of Social-DualCVAE is with nearly $14\%$ less than the best baseline method, Social-STGCNN, and the $FDE_K$ of our model is also with about $8\%$ less. Our method without scene context even outperforms some baselines with scene context, such as Sophie and PIF. We see that Social-DualCVAE with hypatameter 4 outperforms  CVAE alone, implying the learning of social interactions category does help for trajectory forecasting. However, Social-DualCVAE with hyparameter 2 or 6 does not help improve performance. The performance degrading shows that the selection of the inappropriate number of categories will cause misleading classification results, even undermine the prediction performance.  
\subsection{Qualitative Results}
\begin{figure}[ht]
    \centering
    \includegraphics[width=8cm]{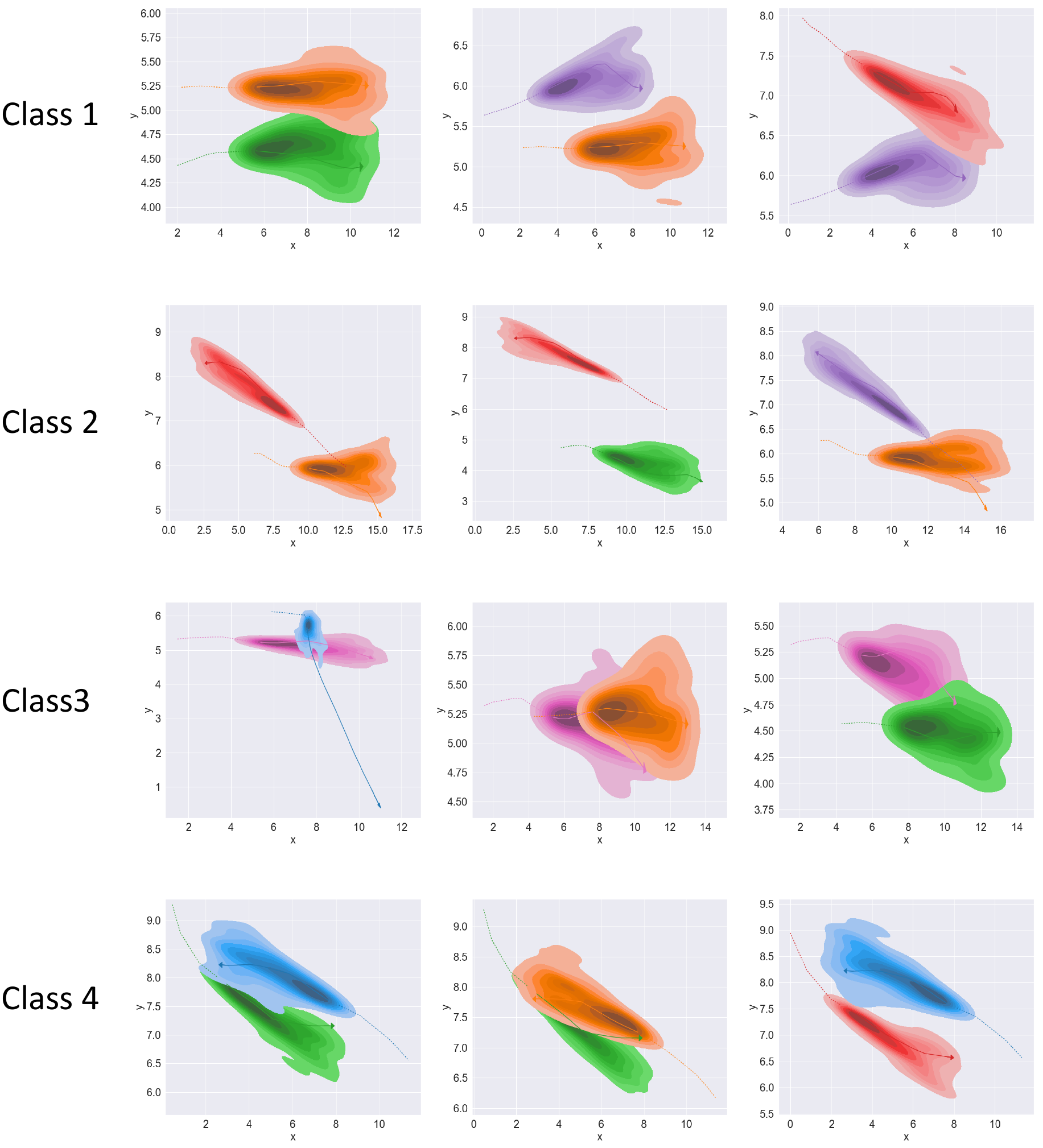}
    \caption{Visualizations of social interactions sampled from the classification results in Zara1 scene, with hyparameter of total categories set to 4. }
    \label{fig5}
\end{figure}
In order to  verify the effectiveness of our proposed pattern aware  method in multi-modal trajectory forecasting more intuitively, we show the generated trajectories under five scenes sampled from ETH and UCY datasets (Fig.~\ref{fig4}). These five scenarios contained usual social interactions in real-world. Compared with Social-STGCNN, Social-DualCVAE could capture the social interactions better. As shown in scene2, the red agent in Social-DualCVAE attempts to avoid collision with the blue one. Social-DualCVAE also shows the smaller variance in multi-modal future trajectories, with more concentrated multi-modal predicted region. Additionally, the multi-modal predicted rigion covers more target trajectory in Social-DualCVAE, represented a better multi-modal prediction accuracy. 

We further visualize the typical social interactions in each category sampled from unsupervised classification results in Zara1 scene (Fig.~\ref{fig5}). In the all social interactions among 38382 pairs of pedestrian combinations instance, 25519 instances belong to class 1, which mainly includes  parallel-walking or parallel-trends in the same direction.  5263 instances belong to class 2. The predicted regions of most trajectories in class 2 do not overlap, meaning a weak interaction or no-interaction between pedestrians. 1854 instances belong to class 3, which mainly contains social interactions of same direction with  catching up or crossing trends. 5746 instances belong to class 4. Most trajectories in the class 4 have the opposite direction with more overlap, representing a certain meeting or collision possibility. Interestingly, Social-DualCVAE achieved unsupervised classification on two plain dimensions, directionality and intersectionality, meanwhile, covers common intricate social interactions.



\section{Conclusion and Future Work}
\label{sec6}
In this article, we presented Social-DualCVAE, a novel architecture for multi-modal pedestrian trajectory forecasting that outperforms prior state-of-the-art methods over several publicly available benchmarks. Different from prior works, our model is able to improve the multimodal predictive performance by digging out the prior knowledge of the social interactions patterns. Through the evaluations and visualizations, we demonstrated that Social-DualCVAE is able to classify the intricate social interaction pattern and generate plausible trajectory distribution for multiple pedestrians. We further analyzed the data flow and developed an effective loss function for DualCVAE training. As shown in experiments, benefiting from unsupervised classification module and latents estimation module, our Social-DualCVAE is able to generate mualtimodal trajectories that more reasonably and realistically predict pedestrian movements. In the future work, we aim to extend the proposed pattern aware model in multi-objects trajectory forecasting, such as urban scenes with different objects, including multiple transportation and pedestrian.

\nobibliography*
\appendix
\section{Reference}
\label{sec:reference_examples}
\bibentry{hb:95}.\\[.2em]
\bibentry{ah:16}.\\[.2em]
\bibentry{bk:16}.\\[.2em]
\bibentry{kt:12}.\\[.2em]
\bibentry{wk:14}.\\[.2em]
\bibentry{pg:10}.\\[.2em]
\bibentry{hc:97}.\\[.2em]
\bibentry{z:19}.\\[.2em]
\bibentry{kp:15}.\\[.2em]
\bibentry{bt:18}.\\[.2em]
\bibentry{l:17}.\\[.2em]
\bibentry{ks:19}.\\[.2em]
\bibentry{mh:20}.\\[.2em]
\bibentry{j:16}.\\[.2em]
\bibentry{kt:21}.\\[.2em]
\bibentry{vs:17}.\\[.2em]
\bibentry{sh:15}.\\[.2em]
\bibentry{li:19}.\\[.2em]
\bibentry{y:21}.\\[.2em]
\bibentry{fr:13}.\\[.2em]
\bibentry{hb:05}.\\[.2em]
\bibentry{gp:18}.\\[.2em]
\bibentry{lc:19}.\\[.2em]
\bibentry{yao:21}.\\[.2em]
\bibentry{pl:09}.\\[.2em]
\bibentry{lr:07}.\\[.2em]
\bibentry{ch:19}.\\[.2em]
\bibentry{sd:19}.\\[.2em]

\nobibliography{aaai22}

\end{document}